\documentclass[letterpaper]{article} 
\usepackage{aaai2026}  
\usepackage{times}  
\usepackage{helvet}  
\usepackage{courier}  
\usepackage[hyphens]{url}  
\usepackage{graphicx} 
\usepackage{natbib}  
\usepackage{caption} 
\usepackage{algorithm}
\usepackage{algorithmic}
\usepackage{amsmath}
\usepackage{amsthm} 
\usepackage{amssymb}
\usepackage{cleveref}
\usepackage{multirow}
\usepackage{booktabs}
\usepackage{tabularx}
\newtheorem{definition}{Definition} 
\newtheorem{assumption}{Assumption}

\usepackage{newfloat}
\usepackage{listings}
\DeclareCaptionStyle{ruled}{labelfont=normalfont,labelsep=colon,strut=off} 
\floatstyle{ruled}
\newfloat{listing}{tb}{lst}{}
\floatname{listing}{Listing}
\title{PITE: Multi-Prototype Alignment for Individual Treatment Effect Estimation}
\author {
    Fuyuan Cao\textsuperscript{\rm 1},
    Jiaxuan Zhang\textsuperscript{\rm 1}\thanks{Corresponding author},
    Xiaoli Li\textsuperscript{\rm 2}
}
\affiliations {
    \textsuperscript{\rm 1}School of Computer and Information Technology, Shanxi University, Taiyuan, China\\
    \textsuperscript{\rm 2}Singapore University of Technology and Design, Singapore\\
    cfy@sxu.edu.cn, 
    jiaxuan7531@163.com, 
    xiaoli\_li@sutd.edu.sg
}

\usepackage{bibentry}

\begin{document}

\maketitle

\begin{abstract}

Estimating Individual Treatment Effects (ITE) from observational data is challenging due to confounding bias. Most studies tackle this bias by balancing distributions globally, but ignore individual heterogeneity and fail to capture the local structure that represents the natural clustering among individuals, which ultimately compromises ITE estimation. While instance-level alignment methods consider heterogeneity, they similarly overlook the local structure information. To address these issues, we propose an end-to-end Multi-\textbf{P}rototype alignment method for \textbf{ITE} estimation (\textbf{PITE}). PITE effectively captures local structure within groups and enforces cross-group alignment, thereby achieving robust ITE estimation. Specifically, we first define prototypes as cluster centroids based on similar individuals under the same treatment. To identify local similarity and the distribution consistency, we perform instance-to-prototype matching to assign individuals to the nearest prototype within groups, and design a multi-prototype alignment strategy to encourage the matched prototypes to be close across treatment arms in the latent space. PITE not only reduces distribution shift through fine-grained, prototype-level alignment, but also preserves the local structures of treated and control groups, which provides meaningful constraints for ITE estimation. Extensive evaluations on benchmark datasets demonstrate that PITE outperforms 13 state-of-the-art methods, achieving more accurate and robust ITE estimation.

\end{abstract}

\begin{links}
    \link{Extended version}{https://aaai.org/example/extended-version}
\end{links}

\section{Introduction}

\begin{figure}[t]
\centering
\includegraphics[width=\linewidth, trim=0 43 0 5, clip]{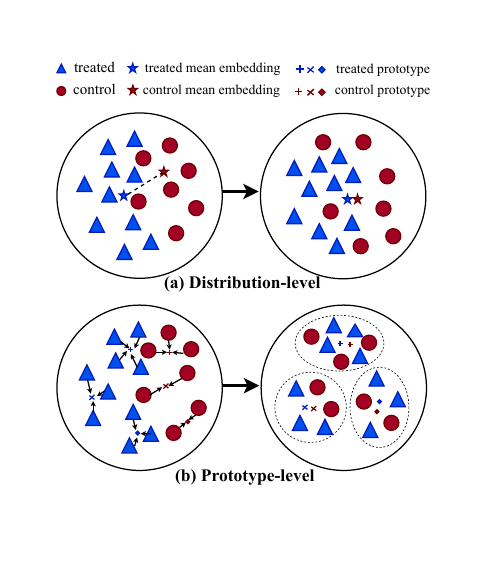}
\caption{(a) Existing distribution-level alignment methods align the overall distributional statistics (e.g., mean) of covariates between treated and control groups but fail to preserve semantic information. (b) Our method achieves both fine-grained prototype-level alignment to reduce distribution shift and preserves local structure between treated and control groups.}
\label{Introduction}
\end{figure}

Estimating the Individual Treatment Effects (ITE) from observational data is critical for personalized decision-making in fields such as healthcare and E-commerce \cite{1,2,3,4}, where understanding the causal impact of interventions guides critical decisions at an individual level. Unlike randomized controlled trials (RCTs), observational studies suffer from \textit{confounding bias} due to confounders \cite{5,6},  variables that influence both treatments and outcomes, which make treated and control groups follow different covariate distributions.

One popular approach for handling confounding bias in treatment effect estimation is distribution-level covariate balance, which aligns the overall distributional statistics of treated and control groups,  as shown in Figure \ref{Introduction} (a). For example, Maximum Mean Discrepancy (MMD) minimizes the distribution discrepancy between treated and control groups by aligning their mean representations. Similarly, adversarial training \cite{10,11} makes factual and counterfactual distributions indistinguishable, naturally mitigating distribution shift. And optimal transport methods \cite{7,8} achieve distribution alignment by moving masses from one distribution to another with minimal transport cost, effectively reducing distribution bias. However, these methods only achieve global distribution balance, neglecting individual-level heterogeneity and the underlying local structures that reflect natural clustering among individuals, which ultimately compromises ITE estimation.

Recent works adapt instance-level alignment methods to address individual heterogeneity, but still neglect local structure constraints during the representation learning process. They generally employ contrastive learning to learn an embedding space where 'positive samples' are pulled closer together and 'negative samples' are pushed apart \cite{x8, zhangcounterfactual}. 
However, the instance-to-instance matching overemphasizes the characteristics of each instance, leading to poor generalization. Meanwhile, these methods also disrupt the natural clustering structure of data during the representation learning process, which degrades the estimation performance.

The local structural information overlooked by both distribution-level and sample-level alignment methods, provides meaningful constraint for ITE estimation. For example, in precision medicine, these methods ignore the inherent patient subgroups during representation learning. Patients typically fall into three response categories  based on drug sensitivity: normal responders, hyper-responders, and low responders \cite{9}. This leads to mismatched pairs where, for instance, a hyper-responder from the treated group might be incorrectly matched with the subgroup of normal responders or low responders from the control group, rather than with the subgroup of hyper-responders. Such prototype-agnostic alignment frequently produces significant errors for hyper-responders, who risk adverse effects.

To overcome these limitations, we propose a Multi-\textbf{P}rototype alignment method for \textbf{ITE} estimation (\textbf{PITE}). PITE effectively captures local structure within groups and enforces cross-group alignment. Specifically, we first define prototypes as cluster centroids based on similar individuals under the same treatment, and then integrate two key techniques: 
(1) Within-group Prototype Matching,
which performs instance-to-prototype matching to assign individuals to the nearest prototype within groups. Instead of global matching, matching a sample to a prototype is more robust to abnormal instances, especially in scenarios with significant individual heterogeneity. 
(2) Cross-group Prototype Alignment, which establishes correspondence between treated and control prototypes to encourage the matched prototypes to be close in the latent space. This dual strategy in PITE enables robust prototype-level alignment, effectively mitigating distribution shift while preserving local structure similarity, thereby making PITE more accurate and robust for instance-level treatment effect predictions. The proposed prototype-level alignment method introduces $k$ prototypes, where the flexible choice of $k$ unifies distribution-level and instance-level alignment. When $k=1$, it performs distribution-level alignment; when $k=n$, it aligns instances. For $1<k<n$, this group-level alignment effectively balances the trade-off between global and individual.


Our main contributions are summarized as follows:

\begin{itemize}
    \item We define prototypes as cluster centroids of similar instances and perform instance-to-prototype matching, thereby capturing the local structure constraints within groups. 
    \item We provide a novel algorithm, PITE, to capture local structure within groups and enforces cross-group alignment for individual treatment effects estimation. 
    \item We conduct a comprehensive evaluation of PITE. Importantly, we find that PITE significantly outperforms distribution-level and instance-level methods, with up to 33.8\% and 39.3\% reduction in estimation error on IHDP, achieving more accurate ITE estimation.

\end{itemize}

\section{Related Work}

Recently, numerous deep learning studies have analyzed the relationship between treatment and outcome at the individual level through mitigating distribution shift, which can be broadly categorized into distribution-level alignment and instance-level alignment methods.

\subsection{Distribution-level alignment}

Current distribution-level alignment methods aim to balance the distributions globally by learning first-order moments, primarily employing distance metrics, adversarial training, and optimal transport techniques. For example, \citet{15} developed TARNet / CFRNet to mitigate confounding bias by reducing the distribution divergence between treated and control groups in the representation space, adopting Maximum Mean Discrepancy (MMD) and Wasserstein distance. GANITE \cite{18} utilized adversarial training to make the discriminator unable to distinguish whether the input data come from the factual distribution or the generate counterfactual distribution. CBRE \cite{k4} introduced an information loop to preserve predictive information that might otherwise be lost during the raw-to-latent space transformation in adversarial training. Alternatively, optimal transport-based methods have also been explored, where \citet{7} reduces the balancing error under the framework of optimal transport with learnable marginal distributions and the cost function. Similarly, \citet{a6} proposed an estimator based on optimal transport to handle both mini-batch sampling effects and unobserved confounder effects issues.

While these methods focus on global distributional alignment, they often neglect the individual-level heterogeneity and intrinsic structure of data such as subgroup similarity or local clustering, which leads to less informative representations and compromises ITE estimation.

\subsection{Instance-level alignment}

Instance-level alignment methods work by matching similar units from different groups to construct locally balanced distributions. The propensity score matching \cite{a1} computes unit similarity based on propensity scores. Instead, representation learning-based methods perform instance-level alignment in learned representation spaces. For example, SITE \cite{16} employed representation learning to capture instance-level variation by selecting specific sample pairs for alignment in the learned embedding space. Similarly, \citet{x8} designed a contrastive task for ITE estimation based on propensity score learning within a representation framework, regarding samples with propensity scores close to 0.5 as positive samples to learn balanced representations. FCCL \cite{zhangcounterfactual} further integrated diffeomorphic counterfactual generation and contrastive learning to address distribution shift through instance-level alignment in the representation space. However, these approaches only achieve partial balance and fail to effectively mitigate the distribution shift.

Compared with instance-level alignment methods, we not only account for individual heterogeneity by performing instance-to-prototype matching that preserves local structural information, but also achieve distributional balance across treatment groups through prototype-level alignment in the latent space, thereby enabling more robust and accurate ITE estimation.

\section{Preliminary}

Following the Neyman-Rubin potential outcome framework \cite{x18,15}, we formally define the problem setup. Let $\mathcal{X} \subset \mathbb{R} ^{d}$ denote the $d$-dimensional covariate space, $\mathcal{T} = \{0, 1\}$ represent the binary treatment space, and $\mathcal{Y} \subset \mathbb{R}$ denote the potential outcome space. We assume the observed dataset contains $n$ independent and identically distributed samples, represented as $\mathcal{D} = \{x_i, t_i, y_i\}_{i=1}^{n}$. For each sample, the covariates are denoted by $x_i \in \mathcal{X}$, and the treatment assignment is defined by the binary variable $t_i \in \mathcal{T}$, where $t_i = 0$ indicates that the $i$-th sample belongs to the control group, and $t_i = 1$ indicates that the $i$-th sample belongs to the treatment group. Each sample has two potential outcomes: $y_i^0$ represents the potential outcome for the $i$-th sample when not receiving treatment, and $y_i^1$ represents the potential outcome for the $i$-th sample when receiving treatment. The actually observed outcome $y_i^{t_i} \in \mathcal{Y}$ only reflects the result under the sample's actual assigned treatment status (\textit{i.e.}, the factual outcome), while the outcome under the unassigned status (the counterfactual outcome $y_i^{1-t_i}$) cannot be directly observed. The observed outcome can be expressed as: $y_i=\left ( 1-t_i\right )y_i^{0} + t_iy_i^{1}$. 

We illustrate with a drug development example that analyzes the efficacy of a newly developed medication for specific patients. In this context, treatment assignment $t_i$ indicates whether a patient received the new medication ($t_i = 1$) or no treatment ($t_i = 0$). The patient's covariates $x_i$ include baseline clinical characteristics such as sex, age, weight, etc. The outcomes $y_i^1$ and $y_i^0$ represent the patient's blood sugar levels with and without the new medication, respectively.

The individual treatment effect (ITE) of sample $i$ is defined as the difference between the potential treatment and control outcomes:
\begin{equation}
\text{ITE}_i =y_{i}^{1} -y_{i}^{0}
\end{equation}

We made the following assumptions to ensure that treatment effects are identifiable:

\begin{assumption}[Consistency]
For a unit with treatment assignment $t$, the observed outcome equals potential outcome $ y^t$.
\end{assumption}

\begin{assumption}[Ignorability]
The potential outcomes are independent of the treatment conditioning on covariates, such that $(y^1,y^0) \perp\!\!\!\perp t|x$. 
\end{assumption}

\begin{assumption}[Overlap]
For any $x$, the probability of receiving treatment is positive. That is, $0< P(t=1|\textit{x})< 1$, for $\forall x\in \mathcal{X}$.
\end{assumption}

\section{Methodology}

\begin{figure*}[ht]
\begin{center}
\includegraphics[width=\linewidth, trim=18 28 0 10, clip]{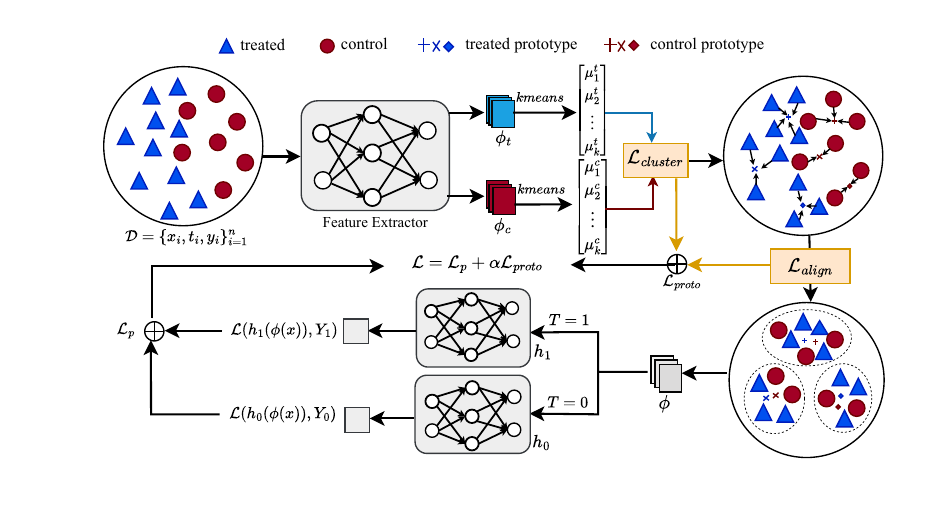}
\caption{An overview of the PITE framework. We perform prototype learning on the representations $\phi_t$ and $\phi_c$ for treated and control groups respectively via k-means to capture local structures within each group. Cross-treatment prototype alignment ($\mathcal{L}_{align}$) enforces correspondence between treated and control prototypes to reduce distribution shift. Finally, two separate neural networks, $h_{1} (\phi(x))$ and $h_{0} (\phi(x))$, are used to estimate potential outcomes under different treatments.}
\label{Research framework}
\end{center}
\end{figure*}

We propose a novel Multi-\textbf{P}rototype Alignment framework for \textbf{I}ndividual \textbf{T}reatment \textbf{E}ffect Estimation (PITE), which integrates three key techniques: (1) Within-group Prototype Matching, which performs instance-to-prototype matching to assign individuals to the nearest prototype; (2) Cross-group Prototype Alignment, which enforces correspondence between matched prototypes across treatment arms; (3) Two-head prediction networks, which predict potential outcomes for treatment and control groups separately based on the learned balanced representations. 
The overall model architecture is presented in Figure \ref{Research framework}.

\subsection{Within-group Prototype Matching}

Prototypes serve as representative embeddings of semantically similar samples \cite{yue2021prototypical,an2024transfer}, providing a stable representation that is less sensitive to individual heterogeneity. Therefore, we first define prototypes and leverage prototypes to identify the natural clustering structures among individuals, thereby reducing bias caused by subgroup differences.

\begin{definition}[Prototype]
A prototype is defined as a learnable cluster centroid that represents a group of individuals with similar hidden representations under the same treatment condition. Formally, for each group $t \in \{0,1\}$, PITE maintains a set of $K$ prototypes:
\begin{equation}
\mu_t = \{\mu_{t,k}\}_{k=1}^K \in \mathbb{R}^{K \times d_h},
\end{equation}
where $d_h$ is the dimension of the hidden representation space. 
\end{definition}

During training, each sample is assigned to its nearest prototype based on Euclidean distance:

\begin{equation}
k^i = \arg\min_{k \in [1,K]} \left\| \phi_i - \mu_{t,k} \right\|_2^2,
\end{equation}
 
\noindent where $k^i$ is the assigned prototype index for sample $i$, $\phi_i$ is the feature representation of sample $i$, and $\mu_{t,k}$ is the $k$-th prototype of group $t$.

We define the clustering loss as:
\begin{equation}
\mathcal{L}_{\text{cluster}} = \sum_{t \in \{0,1\}} \sum_{k=1}^K \sum_{i \in \mathcal{S}_{t,k}} \| \phi_i - \mu_{t,k} \|^2,
\end{equation}

\noindent where \( \phi_i \) denotes the representation of instance \( i \), and \( \mu_{t,k} \) is the prototype for cluster \( k \) in group \( t \). Each sample \( i \) is assigned to a prototype $k^{i}$, and the assignment set is:
\begin{equation}
\mathcal{S}_{t,k} = \{ i \mid k_i^* = k,\; t_i = t \}.
\end{equation}
The gradient with respect to the prototype \( \mu_{t,k} \) is given by:
\begin{equation}
\frac{\partial \mathcal{L}_{\text{cluster}}}{\partial \mu_{t,k}} = \frac{2}{|\mathcal{S}_{t,k}|} \sum_{i \in \mathcal{S}_{t,k}} ( \mu_{t,k} - \phi_i ).
\end{equation}
This objective encourages instance representations to stay close to their corresponding prototypes, thereby preserving the local structure in the representation space.

PITE defines prototypes as learnable cluster centroids of hidden representations within each group, serving as stable and representative anchors for each subgroup. During training, each sample is assigned to the nearest prototype based on Euclidean distance, which ensures clear clustering boundaries and helps identify the natural clustering structures among individuals. The gradient update mechanism ensures that prototypes converge toward the centroids of their assigned samples, obtaining stable and representative cluster centers. Thus, within-group prototype matching provides a stable structural constraint, which are subsequently utilized in cross-group prototype alignment to achieve robust counterfactual estimation.


\subsection{Cross-group Prototype Alignment}

To address the distribution shift between treated and control groups, PITE performs a pairwise prototype alignment strategy through meaningful cross-group prototype matching in the latent space.
Unlike global alignment methods that average over the entire group, we align prototypes—each representing a distinct local cluster, based on the motivation that they capture the local structure in the representation space that represents the natural clustering among individuals. By enforcing proximity between matched prototypes across groups, this loss effectively reduces the distribution mismatch, encourages cross-group correspondence at the prototype-level alignment, and facilitates more reliable counterfactual estimation at a finer granularity. Formally, the alignment objective is defined as: 
\begin{equation}
\mathcal{L}_{\text{align}}= \frac{1}{K} \sum_{k=1}^K \left\| \mu_{1,k} - \mu_{0,k} \right\|_2^2, 
\end{equation}

\noindent where $\mu_{1,k}$ and $\mu_{0,k}$ are the $k$-th prototypes of the treated and control groups, respectively.

However, aggressive alignment may cause prototype collapse. To preserve diversity, PITE introduces a diversity regularization term:

\begin{equation}
\mathcal{L}_{\text{div}} = -\frac{1}{K(K-1)} \sum_{t \in \{0,1\}} \sum_{i \neq j} \left\| \mu_{t,i} - \mu_{t,j} \right\|_2^2,
\end{equation}

\noindent where $t \in \{0,1\}$ denotes the treatment and control groups, $\mu_{t,i}$ represents the $i$-th prototype in group $t$, $K$ is the number of prototypes each group. This regularization encourages each prototype to capture distinct feature patterns within each group. This design balances two key objectives: cross-group prototype alignment, which is essential for accurate individual treatment effect estimation, and intra-group diversity preservation, which prevents information redundancy. By maintaining rich and heterogeneous representations, it ultimately enhances both the accuracy and robustness of causal effect estimation.

The overall prototype loss combines clustering, alignment, and diversity objectives:
\begin{equation}
\mathcal{L}_{\text{proto}} = \mathcal{L}_{\text{cluster}} + \beta \mathcal{L}_{\text{align}} + \gamma \mathcal{L}_{\text{div}},
\end{equation}
where $\beta$ and $\gamma$ are hyperparameters that weight the alignment and diversity terms relative to the clustering objective.

\subsection{Prediction Head}

The learned balanced representations $ \phi(x_i) $ are fed into two neural networks to predict potential outcomes for treatment ($ t=1 $) and control ($ t=0 $) \cite{12,13}. The predicted outcomes are defined as $ T_{\text{out}} = h(\phi(x_i), t_i = 1) $ and $ C_{\text{out}} = h(\phi(x_i), t_i = 0) $, respectively. The predictive loss is given by:
\begin{equation}
{\mathcal{L}}_{p}=\frac{1}{n}\displaystyle\sum_{i=1}^{n}{w}_{i}\cdot \mathcal{L}\left ( h\left ( \phi\left ( {x}_{i}\right ),{t}_{i}\right ),{y}_{i}\right ),
\end{equation}
where $ w_i = \frac{t_i}{2u} + \frac{1-t_i}{2(1-u)} $, and $ u = \frac{1}{n} \sum_{i=1}^n t_i $.

The end-to-end prototype-level alignment method reduces distribution discrepancy across groups while preserving the intrinsic clustering structure of the data by ensuring within-group cohesion and cross-group alignment.The total loss $\mathcal{L}_t$ combines predictive loss, prototype loss, and regularization:
\begin{equation}
\mathcal{L}_{\text{total}} = \mathcal{L}_{\text{p}} + \alpha \mathcal{L}_{\text{proto}} + \lambda {\left \| W\right \|}_{2}, \label{5.12}
\end{equation}

\noindent where $ \alpha $ and $ \beta $ are adjustable hyper-parameters that control the contributions of prototype loss and regularization loss $\left \| \cdot  \right \| _2$ on model weights $W$ to prevent overfitting.

We train our model by minimizing Equation (\ref{5.12}) and provide the detailed multi-prototype alignment strategy for ITE estimation 
in Algorithm 1 in the Appendix.
This formulation ensures that the learned representations achieve cross-group prototype alignment while accurately predicting potential outcomes, ultimately reducing ITE estimation error.


\section{Experiments}

In this section, we evaluate the performance of the proposed multi-prototype alignment for Individual Treatment Effect Estimation (PITE) method with a series of baselines on synthetic, semi-synthetic and real-world datasets.

\subsection{Datasets}

\textbf{Synthetic}: We generate covariates from the multivariate normal distribution $\mathcal{N}\left(\mathbf{0}, \gamma \cdot \sigma^{2} \cdot\left[\rho \mathbf{1}_{p} \mathbf{1}_{p}^{\top}+(1-\rho) \mathbf{I}_{p}\right]\right)$, where the covariance matrix combines an all-ones matrix $\mathbf{1}_{p} \mathbf{1}_{p}^{\top}$ and an identity matrix $\mathbf{I}_{p}$. The scaling parameter $\gamma \in \left \{ 0.4,0.7,1.0,1.2\right \}$ controls the degree of covariate dispersion. We sample 800 units with parameters $p=10$, $\rho=0.2$, $\sigma^2=3$, $\beta_0=[0.2,...,0.2]$, and $\beta_1=[1.2,...,1.2]$. For each $\gamma$, we generate 30 independent datasets, dividing them into training, validation, and test sets with ratios of 63\%, 27\%, and 10\%, respectively. The data generation process is outlined as follows:

\begin{gather*}
\begin{array}{l}
\mathbf{X}_{i} \sim \mathcal{N}\left(\mathbf{0}, \gamma \cdot \sigma^{2} \cdot\left[\rho \mathbf{1}_{p} \mathbf{1}_{p}^{\top}+(1-\rho) \mathbf{I}_{p}\right]\right), \\
T_{i} \mid \mathbf{X}_{i} \sim \operatorname{Bernoulli}\left( 1/({1 + \exp\left( -\mathbf{1}_{p}^{\top} \mathbf{X}_{i} \right)} \right)), \\
Y_{i}^{0}=\boldsymbol{\beta}_{\mathbf{0}} \mathbf{X}_{i}+\xi_{i}, \quad Y_{i}^{1}=\boldsymbol{\beta}_{\mathbf{1}} \mathbf{X}_{i}+\xi_{i}, \quad \xi_{i} \sim \mathcal{N}(0,1).
\end{array}
\end{gather*}

\textbf{Semi-synthetic (IHDP)}. The IHDP dataset, introduced by Hill \cite{x11} based on the Infant Health and Development Program, is a randomized control trail to assess whether there is influence of specialist visit (treatment) on children`s cognitive scores (outcome). Hill excluded a subpopulation with non-white mothers from the treatment group to cause selection bias. The IHDP dataset consists of 747 samples, comprising 139 treated samples and 608 controlled samples. We use the same 100 datasets, following the standard practice in the field. 

\textbf{Real-world (Jobs)}. The Jobs dataset, combined Lalonde and a randomized study, investigated the causal effect of job training (treatment) on income and employment status after training\cite{e8}. This research constructed a binary classification task, where the goal is to predict unemployment using the feature sets. We use the same 10 datasets as used in \cite{15}, comprising 297 treated samples and 2915 controlled samples with train /validation/test splits with ratios 56/24/20.

\begin{table*}[!t]
\begin{center}
\begin{small}
\begin{tabular}{c|cc|cc|cc|cc}
\toprule
\multirow{2}{*}{Method} & \multicolumn{2}{c|}{$\gamma=0.4$}                          & \multicolumn{2}{c|}{$\gamma=0.7$}                          & \multicolumn{2}{c|}{$\gamma=1.0$}                          & \multicolumn{2}{c}{$\gamma=1.2$}                           \\ \cline{2-9} 
                        & $\sqrt{\epsilon _{PEHE}^{within}}$ & $\sqrt{\epsilon _{PEHE}^{out-of}}$ & $\sqrt{\epsilon _{PEHE}^{within}}$ & $\sqrt{\epsilon _{PEHE}^{out-of}}$ & $\sqrt{\epsilon _{PEHE}^{within}}$ & $\sqrt{\epsilon _{PEHE}^{out-of}}$ & $\sqrt{\epsilon _{PEHE}^{within}}$ & $\sqrt{\epsilon _{PEHE}^{out-of}}$ \\ \hline
OLS-1                   & 8.39(0.38)                  & 8.41(0.84)                  & 10.86(0.43)                 & 10.85(1.34)                 & 12.89(0.59)                 & 13.00(1.68)                 & 14.21(0.59)                 & 14.32(1.55)                 \\
OLS-2                   & 5.92(0.27)                  & 5.94(0.60)                  & 7.64(0.30)                  & 7.64(0.96)                  & 9.05(0.40)                  & 9.13(1.18)                  & 9.97(0.41)                  & 10.07(1.11)                 \\
BART                    & 4.04(0.22)                  & 3.40(0.50)                  & 4.70(0.20)                  & 4.17(0.60)                  & 5.36(0.31)                  & 4.86(0.61)                  & 5.81(0.26)                  & 6.14(0.58)                  \\
KNN                     & 4.55(0.33)                  & 5.50(0.62)                  & 6.48(0.37)                  & 7.26(0.75)                  & 7.91(0.38)                  & 9.08(1.27)                  & 8.76(0.46)                  & 10.27(1.07)                 \\
\midrule
CFR-Wass                & 2.48(0.05)         & 2.48(0.06)         & 3.73(0.05)                  & 3.60(0.09)                  & 4.68(0.07)                  & 4.72(0.14)                  & 5.34(0.07)                  & 5.37(0.14)                  \\
CFR-MMD                 & 2.54(0.05)                  & 2.54(0.06)                  & 3.75(0.05)                  & 3.62(0.09)                  & 4.70(0.07)                  & 4.74(0.14)                  & 5.37(0.08)                  & 5.41(0.14)                  \\
GANITE                  & 4.66(0.03)                  & 4.69(0.06)                  & 6.20(0.03)                  & 6.16(0.07)                  & 7.32(0.03)                  & 7.33(0.07)                  & 8.08(0.03)                  & 8.11(0.08)                  \\
ABCEI                   & 2.73(0.03)                  & 2.75(0.06)                  & 3.74(0.04)                  & 3.57(0.09)                  & 4.61(0.05)                  & 4.73(0.13)                  & 5.19(0.06)                  & 5.19(0.12)                  \\
CBRE                    & 2.91(0.03)                  & 2.93(0.05)                  & 4.01(0.04)                  & 3.85(0.08)                  & 4.95(0.05)                  & 5.02(0.12)                  & 5.77(0.06)                  & 5.73(0.13)                  \\ 
DIGNet                    & 3.17(0.09)                  & 3.18(0.11)                  & 4.09(0.10)                  & 3.97(0.13)                  & 5.05(0.10)                  & 5.10(0.17)                  & 5.78(0.09)                  & 5.81(0.15)                  \\ 
SITE                    & 2.68(0.11)                  & 2.69(0.13)                  & 4.17(0.16)                  & 4.25(0.23)                  & 5.98(0.34)                  & 6.01(0.38)                  & 6.19(0.15)                  & 6.21(0.17)                  \\
CITE                    & 2.69(0.04)                  & 2.71(0.07)                  & 3.81(0.06)                  & 3.70(0.10)                  & 4.68(0.06)                  & 4.74(0.14)                  & 5.39(0.07)                  & 5.41(0.14)                  \\
FCCL                      & 2.56(0.04)                  & 2.58(0.06)                & 3.65(0.05)                   & 3.50(0.09)                 & 4.40(0.06)                   & 4.49(0.13)                & 5.10(0.06)                  & 5.12(0.12)        \\ 
\midrule
\textbf{PITE}           & \textbf{2.22(0.09)}                  & \textbf{2.24(0.10) }                 & \textbf{3.42(0.08)}         & \textbf{3.29(0.10)}         & \textbf{4.30(0.08)}         & \textbf{4.35(0.16)}         & \textbf{4.99(0.10)}         & \textbf{5.02(0.15)}         \\
\bottomrule
\end{tabular}
\caption{Experimental results on Synthetic datasets. The best result in each row is highlighted in \textbf{bold}.}
\label{table Synthetic}
\end{small}
\end{center}
\end{table*}

\begin{table}[!t]
\resizebox{\columnwidth}{!}{
\begin{tabular}{c|cccc}
\toprule
Method        & $\sqrt{\epsilon _{PEHE}^{within}}$ & $\epsilon _{ATE}^{within}$ & $\sqrt{\epsilon _{PEHE}^{out-of}}$ & $\epsilon _{ATE}^{out-of}$ \\ \midrule
OLS-1      & 5.83(0.39)                  & 0.73(0.04)                 & 5.91(0.27)                  & 0.95(0.06)                 \\
OLS-2      & 2.42(0.16)                  & 0.14(0.02)                 & 2.55(0.16)                  & 0.31(0.02)                 \\
BART          & 2.13(0.22)                  & 0.24(0.05)                 & 2.32(0.12)                  & 0.35(0.03)                 \\
KNN           & 2.13(0.08)                  & 0.15(0.05)                 & 4.16(0.23)                  & 0.80(0.05)                 \\
\midrule
CFR-Wass   & 0.71(0.04)                  & 0.27(0.03)                 & 0.83(0.08)                  & 0.28(0.03)                 \\
CFR-MMD    & 0.77(0.05)                  & 0.25(0.04)                 & 0.92(0.09)                  & 0.28(0.04)                 \\
GANITE        & 1.92(0.29)                  & 0.43(0.41)                 & 2.43(0.46)                  & 0.49(0.38)                 \\
ABCEI         & 0.79(0.06)                  & 0.12(0.02)                 & 1.00(0.13)                  & 0.15(0.03)                 \\
CBRE        & 0.59(0.06)                  & 0.11(0.02)                 & 0.66(0.07)                  & 0.13(0.02)                 \\
DIGNet        & 0.60(0.04)                  & 0.15(0.02)                 & 0.67(0.07)                  & 0.16(0.02)                 \\
SITE          & 0.84(0.05)                  & 0.30(0.04)                 & 0.98(0.07)                  & 0.32(0.05)                 \\
CITE          & 0.59(0.06)                  & 0.11(0.02)                 & 0.67(0.14)                  & 0.14(0.02)                 \\
FCCL          & 0.53(0.04)                  &  0.09(0.01)                &  0.64(0.07)                 &  0.12(0.02)                \\ 
\midrule
\textbf{PITE} & \textbf{0.51(0.02)}         &  \textbf{0.09(0.01)}       &  \textbf{0.60(0.04)}        &  \textbf{0.11(0.02)}        \\ 
\bottomrule
\end{tabular}
}
\caption{Within-sample and out-of-sample estimation errors for the metrics (\textbf{Lower is better}) on IHDP dataset.}
\label{table IHDP}
\end{table}

\subsection{Metrics}
On IHDP dataset where the true treatment effect for individual is known, we adopt two commonly evaluation metrics, namely the \textit{ Precision in Estimation of Heterogeneous Effect } (${\epsilon }_{PEHE}$) and the \textit{absolute error of Average Treatment Effect} (${\epsilon }_{ATE}$) defined as:
\begin{equation}
{\epsilon }_{PEHE}=\frac{1}{n}\displaystyle\sum_{i=1}^{n}{\left ( \tau ({x}_{i})-\hat{\tau }({x}_{i})\right )}^{2},
\end{equation}
\begin{equation}
{\epsilon }_{ATE}=\left | \hat{ATE}-ATE\right |=\frac{1}{n}\left | \displaystyle\sum_{i=1}^{n}\left ( {\tau }_{i}-{\hat{\tau }}_{i}\right )\right |,
\end{equation}
\noindent where ${\tau }_{i}$ refer to the ground truth treatment effect, $\hat{\tau }_{i}$ is the estimated treatment effect.

On Jobs dataset, we adopt the \textit{policy risk} $\mathcal{R}_{pol}(\pi_{\hat{\tau}})$ and the bias of \textit{Average Treatment Effect on the Treated} prediction ${\epsilon }_{ATT}$.
\begin{equation}
\begin{split}
\mathcal{R}_{\text{pol}}(\pi_{\hat{\tau}}) = 1 - \bigg[ 
& \Pr(\pi_{\hat{\tau}}(x) = 1) \cdot \mathbb{E} \left[ Y_1 \mid \pi_{\hat{\tau}}(x) = 1 \right] \\
+\, & \Pr(\pi_{\hat{\tau}}(x) = 0) \cdot \mathbb{E} \left[ Y_0 \mid \pi_{\hat{\tau}}(x) = 0 \right] 
\bigg],
\end{split}
\end{equation}
\noindent where ${\pi }_{\hat{\tau }}: \mathcal{X}\rightarrow \left \{ 0,1\right \}$ is an policy induced from an ITE estimator $\hat{\tau }\left ( \cdot \right )$ with ${\pi }_{\hat{\tau }}(x)=1$ if $\hat{\tau }(x)> 0$ and $\hat{\tau }(x)=0$ otherwise. 
\begin{align}
\epsilon A T T=\left | \mid\frac{1}{|\mathcal{T} _{1}|} \sum_{i=1}^{|\mathcal{T} _{1}|} y_{i}^{1}-\frac{1}{|\mathcal{T} _{0}|} \sum_{i=1}^{|\mathcal{T} _{0}|} y_{i}^{0} \mid -\mid \frac{1}{|\mathcal{T} _{1}|} \sum_{i=1}^{|\mathcal{T} _{1}|}\left(\hat{y_{i}^{1}}-\hat{y_{i}^{0}}\right)\mid  \right |,
\end{align}

\noindent where $\left | \mathcal{T} _{1} \right | $ and $\left | \mathcal{T} _{0} \right |$ are the number of the units in the treatment and the control groups, respectively.

\subsection{Comparison with Baseline Approaches}

We compare PITE empirically against the following 13 baselines. These approaches can be mainly divided into two categories: traditional methods and deep learning. We further categorize deep learning methods into distribution-level alignment methods and instance-level alignment methods. 

\textbf{Traditional Methods:} Ordinary least square \textbf{(OLS-1)} using treatment as a covariate; \textbf{(OLS-2)}, predicting outcomes separately for each group; Bayesian additive regression trees \textbf{(BART)} leveraging a sum-of-trees structure; K-nearest neighbor \textbf{(KNN)} matching samples using $k$-nearest neighbors. \textbf{Distribution-level alignment:} \textbf{CFR-Wass} \cite{15} and \textbf{CFR-MMD} \cite{15} are two methods using the Wasserstein and MMD metric for counterfactual regression, respectively; \textbf{GANITE} \cite{18} implicitly learns counterfactual distribution using GANs; \textbf{ABCEI} \cite{11} balances distributions using adversarial learning; \textbf{CBRE} \cite{k4} constructs an information loop during adversarial training to minimize information loss; \textbf{DIGNet} \cite{e2} utilizes individual propensity confusion and group distance minimization. \textbf{Instance-level alignment:} \textbf{SITE} \cite{16}, which preserves local similarity in sample representations; \textbf{CITE} \cite{x8} learns representation based on propensity score; \textbf{FCCL} \cite{zhangcounterfactual} integrates diffeomorphic counterfactual generation and contrastive learning to achieve sample-level alignment.

\subsection{Experimental Results}

\begin{figure*}[!t]
\begin{center}
\includegraphics[width=\linewidth, trim=5 0 0 1, clip]{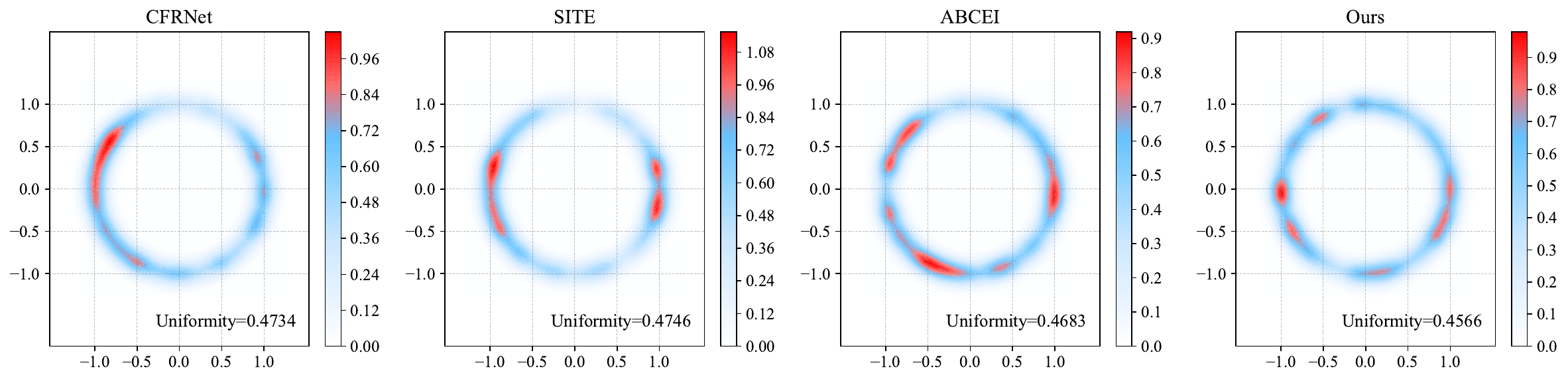}
\caption{Visualization of representation uniformity of four typical methods on IHDP dataset. We visualize the overall feature distributions with Gaussian kernel density estimation (KDE) in ${\mathbb{R}}^{2}$, where the color gradient represents density levels from low (blue) to high (red). The uniformity metric is computed by measuring the pairwise distances between normalized representations on the hypersphere, with lower values indicating superior uniformity.}
\label{uniformity analysis}
\end{center}
\end{figure*}

In this section, we compare and analyse the overall performance of PITE, focusing on robustness under different covariate dispersion conditions. Moreover, we conduct uniformity analysis and sensitivity analysis to validate the efficiency of PITE. Further results, including sensitivity analysis, are presented in the Appendix.

\textbf{Performance Evaluation:} 

We evaluate PITE against baseline methods on the Synthetic, IHDP and Jobs datasets, with the main results shown in Table \ref{table Synthetic} and Table  \ref{table IHDP}, and additional results provided in the Appendix. 

\textbf{Synthetic Data.} Table \ref{table Synthetic} presents the evaluation results of our PITE compared to baseline methods on the synthetic dataset under varying degrees of covariate dispersion ($\gamma = 0.4, 0.7, 1.0, 1.2$). When the covariance parameter $\gamma$ increases from 0.4 to 1.2, PEHE estimation errors universally increase across all methods, indicating that higher data dispersion poses greater challenges for causal effect estimation. Key findings include: (1) PITE consistently achieves the lowest estimation errors across all $\gamma$ settings, significantly outperforming existing methods. (2) Traditional approaches like KNN show dramatic performance degradation as $\gamma$ increases ($\sqrt{\epsilon _{PEHE}^{within}}$ from 4.55 to 8.76), while deep learning methods represented by CFR-MMD also demonstrate poor performance ($\sqrt{\epsilon _{PEHE}^{within}}$ from 2.54 to 5.37). In contrast, PITE maintains remarkable stability with $\sqrt{\epsilon _{PEHE}^{within}}$ increasing only from 2.22 to 4.99. This superior robustness stems from the prototype learning mechanism, which represents natural grouping structures of similar individuals through cluster centroids, thereby avoiding outlier interference inherent in direct global matching and instance-to-instance matching. As data complexity increases, prototypes serve as stable representative points that are more resilient to noise and anomalous observations compared to samples, enabling more accurate and robust ITE estimation.

\textbf{Benchmark Data.} It is worth noting that, PITE significantly outperforms distribution-level alignment and instance-level methods. Compared to \textit{distribution-level} alignment methods such as CFRNet, PITE demonstrates superior performance, achieving substantial reductions in out-of-sample $\epsilon _{PEHE} $ and $\epsilon _{ATE} $ by 34.8\% and 60.7\%, respectively. By performing multi-prototype alignment across groups and preserving the inherent structure during representation learning, PITE effectively mitigates distribution shift and enables more accurate estimation of counterfactual outcomes. CFRNet, GANITE, ABCEI, CBRE and DIGNet show limited performance because these methods generally use the first moment as the distribution discrepancy metric, ignoring the underlying structural constraint that represents the natural clustering among individuals. Compared to \textit{instance-level} alignment methods, PITE outperforms these methods, achieving a 10.4\% reduction in \(\sqrt{\epsilon_{PEHE}^{out-of}}\) compared to CITE. PITE performs instance-to-prototype matching to preserve the local structure in a more robust manner. However, CITE depends heavily on the correct specification of the propensity score, which is usually difficult to obtain. Besides, SITE only achieves partial balance through selecting specific sample pairs for alignment, and therefore shows inferior performance ($\sqrt{\epsilon _{PEHE}^{within}}=0.84$ and $\sqrt{\epsilon _{PEHE}^{out-of}}=0.98$). Although FCCL demonstrates competitive performance compared with distribution-level methods, it suffers from high computational overhead and similarly overlooks local structure preservation, which ultimately compromises ITE estimation. Besides, we evaluate the contribution of the alignment loss and diversity regularization term in the prototype-level alignment in the Appendix.

\textbf{Uniformity Analysis:} 

Figure \ref{uniformity analysis} evaluates the uniformity of four typical methods in the representation space. We observe that our PITE  obtains the lowest uniformity metric $uniformity=0.4566$, which shows that our method can make feature vectors roughly uniformly distributed on the unit hypersphere and preserve as much sample information as possible. PITE assigns instances to semantically meaningful cluster centroids via within-group prototype matching, promoting structured coverage of the representation space within each treatment group. Simultaneously, PITE establishes correspondence between treated and control prototypes through cross-group alignment, preventing isolated clusters and ensuring balanced distribution across treatment arms. By operating on stable cluster representatives rather than instances, PITE provides more robust alignment that effectively prevents representation collapse and achieves more uniform feature space utilization, and enables more accurate ITE estimation.

\section{Conclusion}

In this paper, we address the critical issue of neglecting local structure information that represents the natural clustering among individuals, which exists in both distribution-level and instance-level alignment methods for individual treatment effect estimation. To achieve this, we propose PITE, a novel prototype-level method for robust ITE estimation. PITE innovatively introduces prototypes and designs intra-group instance-to-prototype matching along with cross-group multi-prototype alignment strategies, effectively mitigating distribution shift while preserving the local structure of data, which provides meaningful constraints for ITE estimation. Compared to other baselines, comprehensive experiments across various datasets demonstrate that PITE achieves more accurate and robust ITE estimation. In future work, we will explore causal effect estimation in multimodal data settings, incorporating semantic information across different modalities to enhance ITE estimation performance.

\section*{Acknowledgements}

This work is supported by the National Natural Science Foundation of China (U24A20323, 62376145), the Science and Technology Innovation Talent Team of Shanxi Province (202204051002016), and the Key Technologies Program of Taihang Laboratory in Shanxi Province (THYF-JSZX-24010700).

\bibliography{aaai2026}

\clearpage
\onecolumn
\section*{Appendix}

\section{Additional Technical Details}

All experiments, including the baselines and proposed method, were performed on a consistent hardware configuration featuring an AMD EPYC 7763 64-Core Processor and a single NVIDIA A800 GPU, utilizing TensorFlow 0.12.0-rc1 as our deep learning framework.

PITE integrates three key techniques: (1) Within-group Prototype Matching, which performs instance-to-prototype matching to assign individuals to the nearest prototype; (2) Cross-group Prototype Alignment, which enforces correspondence between matched prototypes across treatment arms; (3) Two-head prediction networks, which predict potential outcomes for treatment and control groups separately based on the learned balanced representations. The end-to-end prototype-level alignment method reduces distribution discrepancy across groups while preserving the intrinsic clustering structure of the data by ensuring within-group cohesion and cross-group alignment.
We provide the detailed multi-prototype alignment strategy for ITE estimation 
in Algorithm 1. This method ensures that the learned representations achieve cross-group prototype alignment while accurately predicting potential outcomes, ultimately reducing ITE estimation error.

\begin{algorithm}[!ht]
\caption{PITE}
\label{Algorithm1}
\begin{algorithmic}[1]
\REQUIRE Training data $\{(\mathbf{x}_i, t_i, y_i)\}_{i=1}^N$, number of prototypes $K$, loss weights $\alpha$, $\beta$, $\gamma$, $\lambda$.
\STATE Initialize encoder $\phi$, predictor $h$, prototypes $\{\mu_{t,k}\}_{k=1}^K$ for $t \in \{0,1\}$.
\STATE Split $\mathcal{D}$ into training set ${\mathcal{D}}_{\text{train}}$ and validation set ${\mathcal{D}}_{\text{valid}}$.
\FOR{each mini-batch in $\mathcal{D}_{\text{train}}$}
    \STATE Sample mini-batch and compute representations $\phi_{t,i} = \phi_t(\mathbf{x}_i)$ for treated samples and $\phi_{c,i} = \phi_c(\mathbf{x}_i)$ for control samples.
    \STATE Assign each sample to nearest prototype $\mu_{t_i,k}$ within groups.
    \STATE Compute clustering loss $\mathcal{L}_{\text{cluster}}$.
    \STATE Compute prototype alignment loss $\mathcal{L}_{\text{align}}$.
    \STATE Compute intra-group diversity loss $\mathcal{L}_{\text{div}}$.
    \STATE Prototype loss: $\mathcal{L}_{\text{proto}} = \mathcal{L}_{\text{cluster}} + \beta \mathcal{L}_{\text{align}} + \gamma \mathcal{L}_{\text{div}}$.
    \STATE Predict outcomes $\hat{y}_i = h(\phi_i, t_i)$ and compute prediction loss $\mathcal{L}_{\text{p}}$.
    \STATE Compute total loss: $\mathcal{L}_{\text{total}} = \mathcal{L}_{\text{p}} + \alpha \mathcal{L}_{\text{proto}} + \lambda \|W\|_2$.
    \STATE Update all parameters until converged on validation set.
\ENDFOR
\STATE \textbf{Return} Trained encoder, predictor, and prototypes.
\end{algorithmic}
\end{algorithm}

We select ELU as the non-linear activation function and adopt Adam optimizer to minimize PITE’s objective function with a learning rate of $1e-3$, and the maximum number of iterations is 3000. Table \ref{para} states the number and range of values tried per hyper-parameter during the training.

\begin{table}[!ht]
\centering
\begin{tabularx}{\textwidth}{>{\centering\arraybackslash}X | >{\centering\arraybackslash}X}
\hline
\textbf{Hyper-parameter}                        & \textbf{Range}                                   \\ \hline
Weight of regularization loss                  & $\{1\text{e}{-3}, 1\text{e}{-4}, 1\text{e}{-5}\}$ \\ \hline
Batch size                                     & $\{64, 128, 256\}$                                \\ \hline
Weight of prototype alignment loss             & $\{10, 20, 30, 40, 50\}$                          \\ \hline
Number of prototypes                           & $\{3, 4, 5, 6\}$                                  \\ \hline
Weight of alignment loss                       & $\{0.1, 0.3, 0.5, 0.7, 0.9\}$                     \\ \hline
Weight of diversity loss                       & $\{1\text{e}{-1}, 1\text{e}{-2}, 1\text{e}{-3}\}$                     \\ 
\hline
\end{tabularx}
\caption{Hyper-parameters and their search ranges}
\label{para}
\end{table}

\section{Additional Experiments on IHDP dataset}

\begin{figure}[!h]
\centering
\includegraphics[width=0.9\linewidth, trim=0 0 0 24, clip]{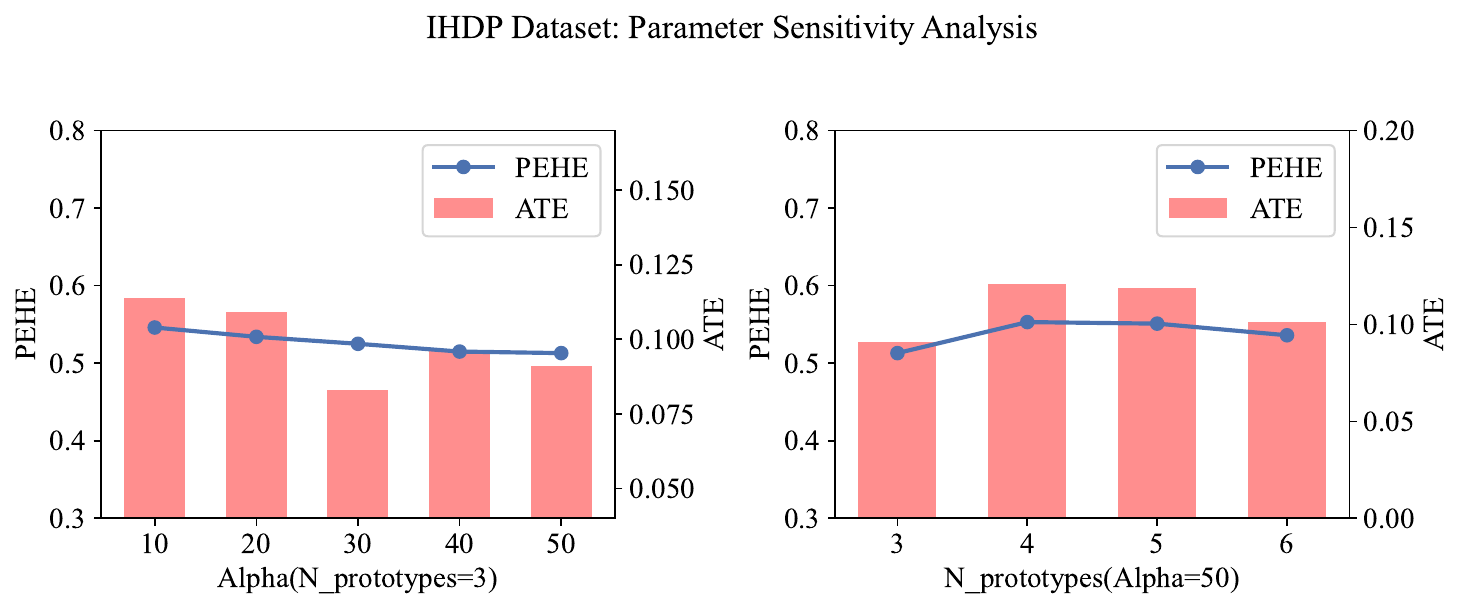}
\caption{ITE estimation performance of PITE under different parameters on IHDP dataset.}
\label{IHDP_sensitivity_analysis}
\end{figure}

We perform sensitivity analysis to examine the robustness of our PITE method by focusing on two key parameters: the weight of prototype alignment loss $\alpha$ and  the number of prototypes $n$. In particular, we set $\alpha \in \left \{ 10,20,30,40,50\right \}$, $n \in \{3,4,5,6\}$. The results in Figure \ref{IHDP_sensitivity_analysis} (a) show that the prototype alignment loss impacts the ITE estimation overall performance, where the dual y-axes represent the $\epsilon _{PEHE}$ and $\epsilon _{ATE}$ metrics, respectively. We find that PITE is generally robust to different settings of $\alpha$. Besides, we observe that the model achieves optimal performance when the number of prototypes $n=3$, and thus we set $n=3$ in our experiments.

\subsection{Ablation Study}

\begin{table}[!t]
\centering
\begin{tabular}{c|cccc}
\toprule
Method        & $\sqrt{\epsilon _{PEHE}^{within}}$ & $\epsilon _{ATE}^{within}$ & $\sqrt{\epsilon _{PEHE}^{out-of}}$ & $\epsilon _{ATE}^{out-of}$ \\ 
\midrule
w/o alignment        & 0.57(0.05)                  & 0.10(0.01)                 & 0.68(0.06)                  & 0.12(0.02)                 \\
w/o  diversity        & 0.59(0.13)                  & 0.09(0.02)                 & 0.72(0.14)                  & 0.11(0.02)                 \\
\textbf{PITE} & \textbf{0.51(0.02)}         &  \textbf{0.09(0.01)}       &  \textbf{0.60(0.04)}        &  \textbf{0.11(0.02)}           \\ 
\bottomrule
\end{tabular}
\caption{Ablation study to the effect of each component on IHDP dataset.}
\label{table ablation}
\end{table}

In Table \ref{table ablation}, we evaluate the contribution of the alignment loss and diversity regularization term in the Cross-group Prototype Alignment. emoving the alignment loss (i.e., w/o alignment) leads to a notable drop in performance, with $\epsilon _{PEHE}^{within}$ increasing to 0.57. Similarly, removing the diversity regularization term (i.e., w/o diversity) results in a further increase in $\epsilon _{PEHE}^{within}$, reaching 0.59. The full model (i.e., PITE) achieves the best performance across all metrics, with $\epsilon _{PEHE}^{within}$ of 0.51 and $\epsilon _{ATE}^{within} $ of 0.09. These results highlight the critical role of both the alignment loss and the diversity regularization term in improving the model’s performance.

\section{Additional Experiments on Synthetic dataset}

\begin{figure}[!htbp]
\begin{center}
\centerline{\includegraphics[width=0.55\columnwidth]{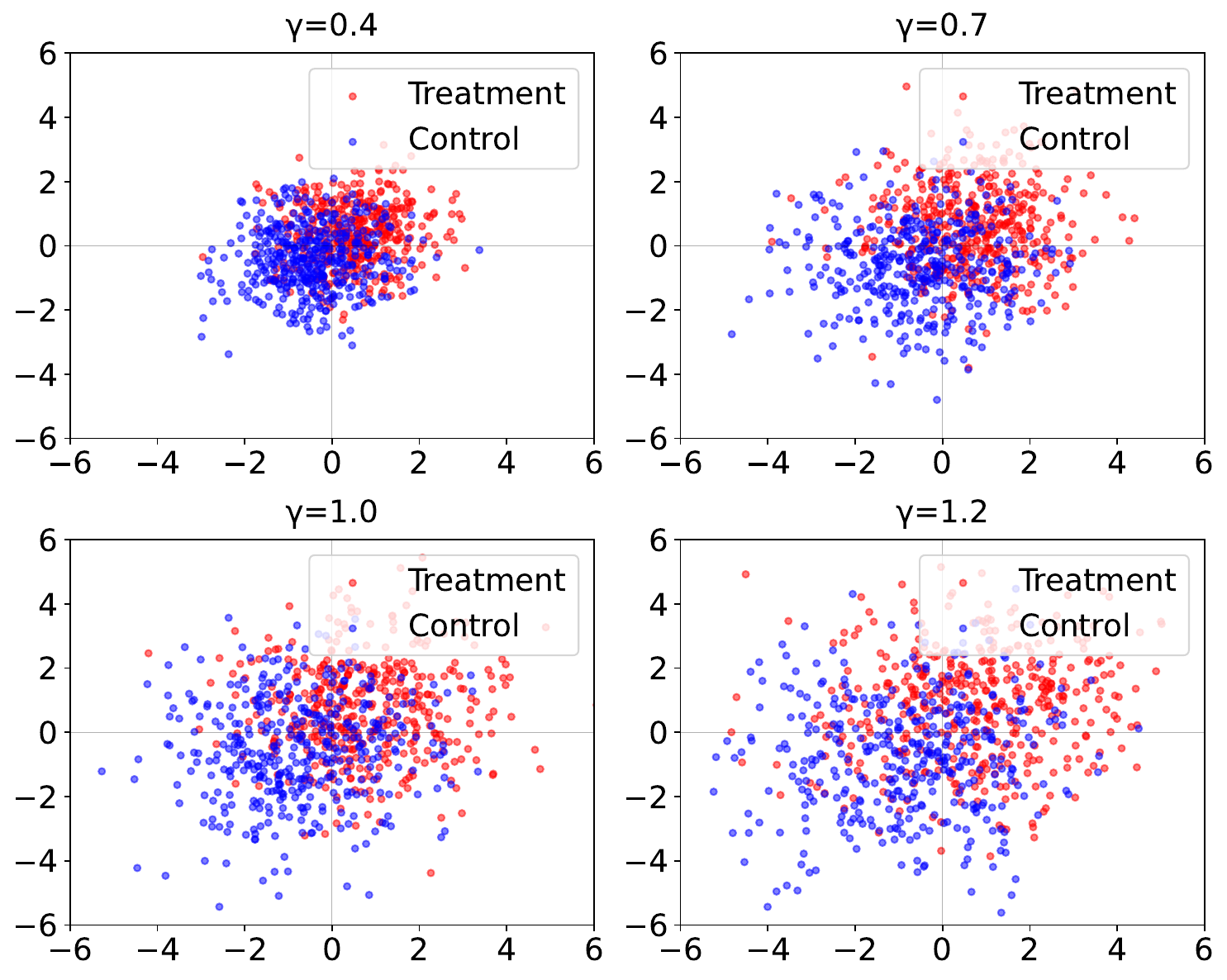}}
\caption{T-SNE visualizations of the covariates as $\gamma$ varies.}
\label{gama varying}
\end{center}
\end{figure}

Following \cite{zhangcounterfactual}, we generate covariates from a multivariate normal distribution $\mathcal{N}\left(\mathbf{0}, \gamma \cdot \sigma^{2} \cdot\left[\rho \mathbf{1}_{p} \mathbf{1}_{p}^{\top}+(1-\rho) \mathbf{I}_{p}\right]\right)$, where the scaling parameter $\gamma \in \left \{ 0.4,0.7,1.0,1.2\right \}$ controls the degree of covariate dispersion, as shown in \cref{gama varying}.

\cref{gama varying} presents T-SNE visualizations of the covariates under different values of $\gamma$. At $\gamma=0.4$, the treated and control groups exhibit relatively concentrated distributions. As $\gamma$ increases to 1.0 and 1.2, the covariate distributions become significantly more dispersed, making the dataset considerably more complex and posing severe challenges for the individual treatment effect estimation. This trend highlights how larger $\gamma$ values represent greater heterogeneity, providing various scenarios for the individual treatment effect estimation. \cref{table Synthetic} and \cref{table Synthetic2} present the ITE estimation error metrics on synthetic datasets.

\begin{table}[!ht]
\begin{center}
\begin{tabular}{c|cccc|cccc}
\toprule
\multirow{2}{*}{Method} & \multicolumn{4}{c|}{$\gamma=0.4$}                          & \multicolumn{4}{c}{$\gamma=0.7$}                                             \\ \cline{2-9} 
                        & $\sqrt{\epsilon _{PEHE}^{within}}$ & $\epsilon _{ATE}^{within}$ & $\sqrt{\epsilon _{PEHE}^{out-of}}$ & $\epsilon _{ATE}^{out-of}$ & $\sqrt{\epsilon _{PEHE}^{within}}$ & $\epsilon _{ATE}^{within}$ & $\sqrt{\epsilon _{PEHE}^{out-of}}$ & $\epsilon _{ATE}^{out-of}$ \\ \hline
OLS-1                   & 8.39(0.38)                  & 4.54(0.30)                  & 8.41(0.84)                 & 4.56(0.56)                 & 10.86(0.43)                 & 5.94(0.34)                 & 10.85(1.34)                 & 6.15(1.06)                 \\
OLS-2                   & 5.92(0.27)                  & 3.18(0.22)                  & 5.94(0.60)                  & 3.19(0.50)                  & 7.64(0.30)                  & 4.16(0.24)                  & 7.64(0.96)                  & 4.32(0.77)                 \\
BART                    & 4.04(0.22)                  & 1.94(0.18)                  & 3.40(0.50)                  & 2.57(0.60)                  & 4.70(0.20)                  & 2.86(0.61)                  & 4.17(0.60)                  & 3.14(0.58)                  \\
KNN                     & 4.55(0.33)                  & 2.80(0.27)                  & 5.50(0.62)                  & 3.57(0.59)                  & 6.48(0.37)                  & 4.25(0.29)                  & 7.26(0.75)                  & 5.03(0.70)                 \\
\midrule
CFR-Wass                & 2.48(0.05)         & 1.12(0.04)         & 2.48(0.06)                  & 1.16(0.09)                  & 3.73(0.05)                  & 1.67(0.14)                  & 3.60(0.09)                  & 1.72(0.09)                  \\
CFR-MMD                 & 2.54(0.05)         & 1.15(0.05)                  & 2.54(0.06)                  & 1.18(0.06)                  & 3.75(0.05)                  & 1.70(0.10)                  & 3.62(0.09)     
& 1.73(0.10)
\\
SITE                    & 2.68(0.11)                  & 1.16(0.12)                  & 2.69(0.13)                  & 1.21(0.12)                  & 4.17(0.16)                  & 1.92(0.09)                  & 4.25(0.23)                  & 1.95(0.13)                  \\
CITE                    & 2.69(0.04)                  & 1.17(0.06)                  & 2.71(0.07)                  & 1.18(0.07)                  & 3.81(0.06)                  & 1.75(0.07)                  & 3.70(0.10)                  & 1.76(0.08)                  \\
GANITE                  & 4.66(0.03)                  & 1.00(0.06)                  & 4.69(0.06)                  & 1.09(0.08)                  & 6.20(0.03)                  & 1.54(0.09)                  & 6.16(0.07)                  & 1.50(0.12)                  \\
ABCEI                   & 2.73(0.03)                  & 1.42(0.04)                  & 2.75(0.06)                  & 1.53(0.07)                  & 3.74(0.04)                  & 1.66(0.04)                  & 3.57(0.09)                  & 1.72(0.08)                  \\
CBRE                    & 2.91(0.03)                  & 1.64(0.07)                  & 2.93(0.05)                  & 1.71(0.07)                  & 4.01(0.04)                  & 2.30(0.10)                  & 3.85(0.08)                  & 2.33(0.11)                  \\ 
DIGNet                    & 3.17(0.09)                  & 1.97(0.07)                  & 3.18(0.11)                  & 1.99(0.08)                  & 4.09(0.10)                  & 2.52(0.07)                  & 3.97(0.13)                  & 2.51(0.08)                  \\ 
FCCL           & 2.56(0.04)                  & 1.05(0.06)                  & 2.59(0.06)         & 1.10(0.07)         & 3.65(0.05)         & \textbf{1.18(0.04)}         & 3.50(0.09)         & \textbf{1.19(0.09)}         \\
\midrule
\textbf{PITE}           & \textbf{2.21(0.10)}                  & \textbf{0.94(0.06)}                  & \textbf{2.24(0.11)}         & \textbf{0.98(0.07)}         & \textbf{3.42(0.08)}         & 1.44(0.07)         & \textbf{3.28(0.10)}         & 1.42(0.09)         \\ 
\bottomrule
\end{tabular}
\caption{Additional experimental results on Synthetic datasets ($\gamma=0.4$ and $\gamma=0.7$).}
\label{table Synthetic}
\end{center}
\end{table}

\begin{table}[!ht]
\begin{center}
\begin{tabular}{c|cccc|cccc}
\toprule
\multirow{2}{*}{Method} & \multicolumn{4}{c|}{$\gamma=1.0$}                          & \multicolumn{4}{c}{$\gamma=1.2$}                                             \\ \cline{2-9} 
                        & $\sqrt{\epsilon _{PEHE}^{within}}$ & $\epsilon _{ATE}^{within}$ & $\sqrt{\epsilon _{PEHE}^{out-of}}$ & $\epsilon _{ATE}^{out-of}$ & $\sqrt{\epsilon _{PEHE}^{within}}$ & $\epsilon _{ATE}^{within}$ & $\sqrt{\epsilon _{PEHE}^{out-of}}$ & $\epsilon _{ATE}^{out-of}$ \\ \hline
OLS-1                   & 12.89(0.59)                  & 7.12(0.44)                  & 13.00(1.68)                 & 7.17(1.33)                 & 14.21(0.59)                 & 7.85(0.47)                 & 14.32(1.55)                 & 8.16(1.27)                 \\
OLS-2                   & 9.05(0.40)                  & 4.99(0.31)                  & 9.13(1.18)                  & 5.01(0.95)                  & 9.97(0.41)                  & 5.49(0.34)                  & 10.07(1.11)                  & 5.71(0.90)                 \\
BART                    & 5.36(0.31)                  & 3.40(0.50)                  & 4.86(0.61)                  & 4.17(0.60)                  & 5.81(0.26)                  & 4.86(0.61)                  & 6.14(0.58)                  & 6.14(0.58)                  \\
KNN                     & 7.91(0.38)                  & 5.28(0.31)                  & 9.08(1.27)                  & 6.22(1.07)                  & 8.76(0.46)                  & 5.89(0.38)                  & 10.27(1.07)                  & 7.30(1.02)                 \\
\midrule
CFR-Wass                & 4.68(0.07)         & 3.26(0.06)         & 4.72(0.14)                  & 3.28(0.09)                  & 5.34(0.07)                  & 2.70(0.14)                  & 5.37(0.14)                  & 2.72(0.14)                  \\
CFR-MMD                 & 4.70(0.07)                  & 2.74(0.06)                  & 4.74(0.14)                  & 2.76(0.09)                  & 5.37(0.08)                  & 2.69(0.14)                  & 5.41(0.14)                  & 2.73(0.14)                  \\
SITE                    & 5.98(0.34)                  & 3.37(0.21)                  & 6.01(0.38)                  & 3.43(0.23)                  & 6.19(0.15)                  & 3.38(0.11)                  & 6.21(0.17)                  & 3.47(0.14)                  \\
CITE                    & 4.68(0.06)                  & 2.32(0.06)                  & 4.74(0.14)                  & 2.36(0.10)                  & 5.39(0.07)                  & 2.62(0.07)                  & 5.41(0.14)                  & 2.74(0.11)                  \\
GANITE                  & 7.32(0.03)                  & 2.52(0.06)                  & 7.33(0.07)                  & 2.56(0.07)                  & 8.08(0.03)                  & 3.23(0.07)                  & 8.11(0.08)                  & 3.28(0.08)                  \\
ABCEI                   & 4.61(0.05)                  & 2.04(0.05)                  & 4.73(0.13)                  & 2.08(0.11)                  & 5.19(0.06)                  & 2.35(0.07)                  & 5.19(0.06)                  & 2.40(0.12)                  \\
CBRE                    & 4.95(0.05)                  & 2.64(0.12)                  & 5.02(0.12)                  & 2.68(0.11)                  & 5.76(0.06)                  & 3.53(0.16)                  & 5.73(0.13)                  & 3.68(0.17)                  \\ 
DIGNet                    & 5.05(0.10)                  & 2.58(0.11)                  & 5.10(0.17)                  & 2.60(0.13)                  & 5.78(0.09)                  & 3.39(0.06)                  & 5.81(0.15)                  & 3.45(0.12)                  \\ 
FCCL           & 4.40(0.06)                  & 2.02(0.06)                  & 4.49(0.13)         & 2.06(0.10)         & 5.10(0.06)         & 2.57(0.08)         & 5.12(0.12)         & 2.61(0.12)         \\
\midrule
\textbf{PITE}           & \textbf{4.30(0.08)}                  & \textbf{1.99(0.06)}                  & \textbf{4.35(0.16)}         & \textbf{2.04(0.17)}         & \textbf{4.99(0.10)}         & \textbf{2.33(0.06)}         & \textbf{5.03(0.15)}         & \textbf{2.40(0.10)}         \\ 
\bottomrule
\end{tabular}
\end{center}
\caption{Additional experimental results on Synthetic datasets ($\gamma=1.0$ and $\gamma=1.2$).}
\label{table Synthetic2}
\end{table}

\section{Additional Experiments on Jobs dataset}

\begin{figure}[!ht]
\begin{center}
\centerline{\includegraphics[width=0.9\linewidth, trim=0 0 0 24, clip]{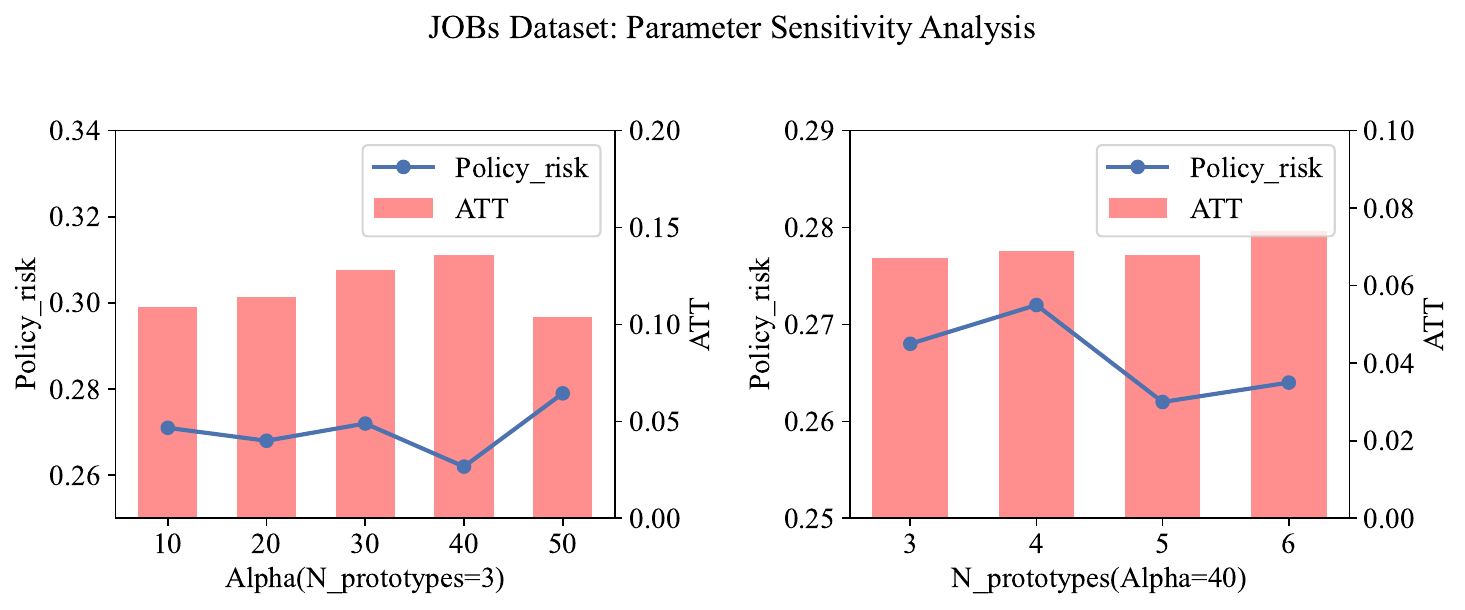}}
\caption{ITE estimation performance of our PITE under different parameters on Jobs dataset.}
\label{jobs_sensitivity analysis}
\end{center}
\end{figure}

On the Jobs dataset, although PITE does not achieve the best performance, it still achieves comparable results to the baselines, particularly in the error metric \(\epsilon_{ATT}\). We also provide sensitivity analysis on Jobs dataset focusing on the weight of prototype alignment loss $\alpha$ and  the number of prototypes $n$.
In particular, we vary the weight of the prototype alignment loss $\alpha \in \left \{ 10,20,30,40,50\right \}$ while fixing $n = 3$. Similarly, we explore the effect of the number of prototypes $n \in \{3,4,5,6\}$ with $\alpha = 40$. 
From the results in \cref{jobs_sensitivity analysis}, we find our model is generally robust to different $\alpha$ settings and the best performance is achieved with $\alpha = 40$.

\begin{table}[!ht]
\begin{center}
\begin{tabular}{ccccc}
\toprule
Method        & $R_{pol}^{within}$  & $\epsilon _{ATT}^{within}$ & $R_{pol}^{out-of}$  & $\epsilon _{ATT}^{out-of}$ \\ 
\midrule
OLS-1      & $0.22(0.02)$          & $0.01(0.00)$                 & $0.23(0.02)$          & $0.08(0.04)$                 \\
OLS-2      & $0.21(0.01)$          & $0.01(0.01)$        & $0.24(0.03)$          & $0.08(0.03)$                 \\
BART         & $0.23(0.00)$         & $0.02(0.02)$                 & $0.25(0.02)$          & $0.08(0.04)$                 \\
KNN           & $0.23(0.01)$         & $0.02(0.00)$                 & $0.26(0.02)$          & $0.13(0.05)$                 \\
\midrule
CFR-Wass   & $0.23(0.01)$          & $0.06(0.02)$                 & $0.26(0.02)$          & $0.10(0.04)$                 \\
CFR-MMD    & $0.22(0.00)$          & $0.07(0.03)$                 & $0.27(0.01)$          & $0.12(0.05)$                 \\
SITE          & $0.23(0.01)$          & $0.05(0.02)$                 & $0.25(0.02)$          & $0.10(0.04)$                 \\
CITE          & $0.23(0.00)$          & $0.10(0.03)$                 & $0.26(0.02)$          & $0.13(0.05)$                 \\
GANITE        & $\boldsymbol{0.14(0.02)}$          & $0.27(0.74)$                 & $\boldsymbol{0.15(0.01)}$ & $0.31(0.56)$     \\
ABCEI         & $0.17(0.02)$          & $0.05(0.02)$                 & $0.22(0.02)$          & $0.15(0.08)$        \\
CBRE        & $0.30(0.00)$          & $0.08(0.03)$                 & $0.31(0.00)$          & $0.11(0.03)$           \\
DIGNet        & $0.23(0.01)$          & $0.07(0.04)$                 & $0.26(0.01)$          & $0.13(0.04)$     \\
FCCL &  $0.23(0.01)$           & $0.05(0.01)$                 & $0.25(0.02)$          & $\boldsymbol{0.07(0.03)}$                 \\ 
\midrule
\textbf{PITE} &  $0.22(0.01)$           & $\boldsymbol{0.04(0.01)}$                 & $0.24(0.02)$          & $0.08(0.02)$                 \\ 
\bottomrule
\end{tabular}
\end{center}
\caption{Within-sample and out-of-sample policy risk and error on the average treatment effect on the treated (ATT) for the various models on Jobs dataset.}
\label{table jobs}
\end{table}

\end{document}